\documentclass[lettersize,journal]{IEEEtran}
\usepackage{amsmath,amssymb,amsfonts}
\usepackage{array}

\usepackage{textcomp}
\usepackage{stfloats}
\usepackage{url}
\usepackage{verbatim}
\usepackage{graphicx}
\usepackage{subcaption}
\def\BibTeX{{\rm B\kern-.05em{\sc i\kern-.025em b}\kern-.08em
    T\kern-.1667em\lower.7ex\hbox{E}\kern-.125emX}}
\usepackage{balance}
\usepackage{algorithm}
\usepackage{algpseudocode}
\usepackage[caption=false,font=normalsize]{subfig}
\usepackage{xcolor}
\usepackage{booktabs}

\usepackage{threeparttable}
\usepackage{multirow}
\newcommand{\Method}{BAT}

\newcommand{\rev}[1]{\textcolor{black}{#1}}

\begin{document}
\title{\Method{}: Balancing Agility and Stability via Online Policy Switching for Long-Horizon Whole-Body Humanoid Control}
\author{Donghoon Baek$^{1}$, Sang-Hun Kim$^{2}$, and Sehoon Ha$^{1}$
\thanks{$^{1}$Georgia Institute of Technology, Atlanta, GA, 30308, USA}%
\thanks{$^{2}$Samsung Research}%
\thanks{*Correspondence to dbaek36@gatech.edu}
}

\maketitle

\begin{abstract}

Despite advances in reinforcement and imitation learning, achieving both precise control and dynamic whole-body behaviors in long-horizon tasks remains challenging. Existing approaches typically follow two paradigms: coupled whole-body policies for global coordination and decoupled policies for modular precision. However, effectively combining these two types of controllers to achieve agility, robustness, and precision remains challenging. In this work, we propose \Method{}, an online policy-switching framework that
dynamically selects between coupled and decoupled whole-body RL generalist controllers according to the evolving motion context. \rev{\Method{} employs two complementary switching predictors: OpVQ-VAE provides motion-token-based predictions with strong generalization, while OpHRL provides closed-loop, state-aware predictions.
A Token-Familiarity Router (TFR) selects between their predictions, favoring OpHRL for familiar token sequences and OpVQ-VAE for unfamiliar ones when their predictions disagree. \Method{} achieves 85.2\% success on seen long-horizon motion combinations, while retaining 71.3\% success on unseen motions.
\Method{} also outperforms existing whole-body controllers on individual motions and demonstrates successful zero-shot deployment on the Unitree G1 humanoid.}

\end{abstract}


\section{Introduction}
\label{introduction}

\rev{Learning-based whole-body humanoid control can be broadly categorized into two paradigms. 
\textit{Decoupled policies} separate upper-body manipulation from lower-body locomotion, enabling precise upper-body tracking and robust locomotion through specialized control objectives~\cite{zhang2025falcon,lu2025mobile}.
In contrast, \textit{coupled policies} use unified motion imitation to enable agile, coordinated whole-body behaviors~\cite{chen2025gmt,luo2025sonic,li2025bfm,ze2025twist}. 
Their complementary strengths are particularly relevant to long-horizon tasks, where control demands shift between precise upper-body execution and dynamic whole-body coordination.}

\begin{figure}[!t]
    \centering
    \includegraphics[width=\columnwidth]{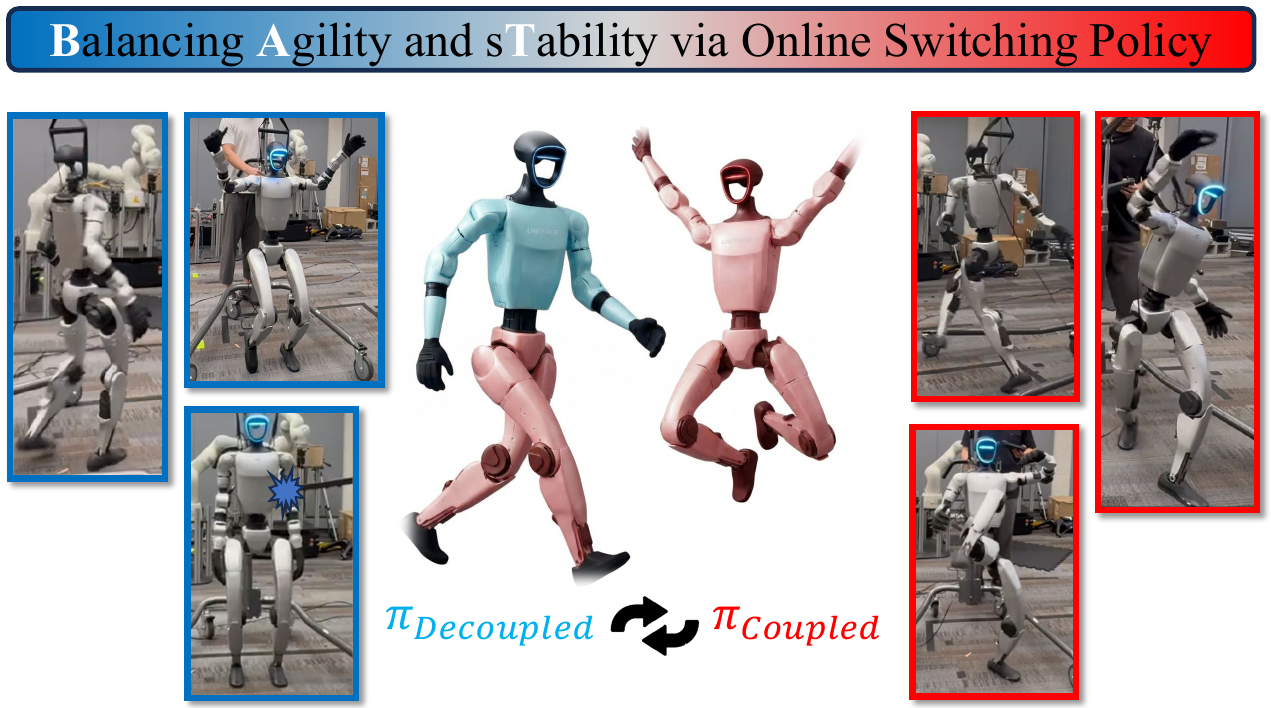}
    \caption{
\textbf{Conceptual overview of BAT}. The decoupled policy provides stability and robustness, while the coupled policy enables agile whole-body motion. BAT adaptively switches between them to combine both strengths.
}
\label{fig1}
\vspace{-1.5
em}
\end{figure}

\rev{These complementary strengths motivate combining decoupled and coupled control within a single long-horizon execution. Both controllers are generalists that support broad motion repertoires, but their different control formulations make them suitable for different motion regimes. We propose \Method{} (Balancing Agility and Stability), an online policy-switching framework that exploits this complementarity by selecting between two fixed generalist controllers as control demands evolve (Fig.~\ref{fig1}). By combining the decoupled policy’s stability and precision with the coupled policy’s agility, \Method{} enables both capabilities in long-horizon, multi-task execution.}

\rev{Realizing this approach across diverse, long-horizon tasks presents two main challenges: (1) the number of possible motion compositions grows combinatorially with the motion repertoire and task horizon, making exhaustive training coverage impractical; and (2) demonstrations provide neither controller-selection nor switch-timing labels, while the consequences of a controller choice may be delayed and depend on the evolving robot state, making reward-only switching difficult to optimize.}

\rev{\Method{} addresses these challenges through two complementary switching predictors. To address challenge (1), an option-aware vector-quantized variational autoencoder (OpVQ-VAE) learns a reusable vocabulary of discrete motion tokens that encode controller preference. A transition-aware option head uses this shared vocabulary across motion compositions, reducing the need for exhaustive training coverage. To address challenge (2), option-guided hierarchical reinforcement learning (OpHRL) uses offline cross-simulator switch-boundary evaluation to obtain controller-selection and timing supervision. This guidance initializes and regularizes the switching policy, reducing reliance on delayed rewards alone, while closed-loop reinforcement learning refines its decisions using proprioceptive feedback and motion references. A Token-Familiarity Router (TFR) integrates the two predictors, favoring OpHRL for familiar token sequences and OpVQ-VAE for unfamiliar ones when their predictions disagree.}

\rev{Our contributions are threefold: (1) an online policy-switching framework that combines the complementary strengths of decoupled and coupled generalist policies to balance stability, precision, and agility in humanoid whole-body control; (2) a switching architecture that combines OpVQ-VAE for motion-based generalization through a reusable, option-aware
token vocabulary with OpHRL for state-aware switching learned through cross-simulator supervision and closed-loop reinforcement
learning, using a Token-Familiarity Router to select between their predictions; and (3) extensive simulation evaluations against diverse baselines, qualitative hardware demonstrations, and an analysis of the complementary characteristics of decoupled and coupled policies.}
\section{Related Work}
\label{Related Work}

\subsection{Whole-Body Humanoid Control}
Model-based approaches have been widely used for whole-body humanoid control \cite{baek2025whole,khazoom2024tailoring}, but their reliance on accurate dynamics and contact models limits robustness. More recently, advances in human demonstrations \cite{mahmood2019amass}, imitation learning \cite{peng2018deepmimic}, reinforcement learning, and large-scale simulation \cite{makoviychuk2021isaac} have enabled learning-based approaches that alleviate these limitations. Building on the strong performance of reinforcement learning for motion tracking in physics-based character animation \cite{peng2018deepmimic}, more recent studies have demonstrated significant progress in transferring policies learned in simulation to real-world robotic systems \cite{chen2025gmt,li2025bfm,luo2025sonic}. \rev{Rather than replacing unified controllers, we extend their capabilities through a motion-aware policy-switching mechanism that allows each controller to operate within its most effective regime, striking a balance between precise manipulation and agile motion dynamics across diverse tasks.}

\subsection{Hybrid and Compositional Approaches to Robot Control}
Effective robot control often requires leveraging multiple specialized controllers that perform well in different regimes. Value-function based switching provides a simple selection mechanism \cite{he2024agile}, but becomes unstable when task boundaries are ambiguous. Mixture-of-experts methods combine controllers via gating \cite{wang2025more}, yet struggle when expert behaviors differ significantly. Meta-learning aims to adapt a unified policy across tasks \cite{nam2022skill}, but may lack sufficient specialization. \rev{Rule- or event-based switching relies on predefined transition conditions, while supervised switching requires expert selection labels, limiting their applicability to unseen motion sequences~\cite{Colledanchise2018BT,Zhang2016SafeDAgger}. Hierarchical reinforcement learning enables temporal abstraction through task decomposition and option selection~\cite{Dietterich2000MAXQ,Bacon2017OptionCritic}, including goal-conditioned hierarchies with off-policy correction~\cite{Nachum2018HIRO}. Prior hierarchical locomotion methods learn reusable primitives~\cite{Li2020ICRA}, coordinate gait-specific policies~\cite{yu2026discovery}, use reference-based skill graphs~\cite{Lau2026Switch}, or enable gait transitions through symmetry-guided hierarchies~\cite{Ding2026Arxiv}. While these approaches organize skills, subgoals, or gaits, \Method{} switches online between fixed generalist whole-body controllers to combine the stability and precision favored by decoupled control with the dynamic coordination favored by coupled control across diverse, long-horizon sequential motion tasks.}

\subsection{Representation Learning}

Compact latent representations provide effective interfaces for learning control in high-dimensional motion spaces. VAE-based methods learn continuous motion manifolds that can serve as action representations for downstream control~\cite{ling2020character},
while recent humanoid controllers encode motions, goals, and behaviors in latent spaces to improve learning and generalization ~\cite{li2025bfm,luo2025sonic}. In contrast, vector-quantized models such as VQ-VAE represent diverse motions using a finite vocabulary of
reusable discrete tokens~\cite{zhang2023generating,guo2024momask}.
Token-based representations have also been explored for sequential humanoid control, where discretized sensorimotor trajectories provide a compact interface for behavior prediction
~\cite{radosavovic2024humanoid}. Inspired by the reuse and recombination of discrete tokens across motions, we employ a shared codebook to represent diverse motions and their long-horizon compositions. Unlike prior work focused primarily on motion generation, reconstruction, or general-purpose control, our option-aware representation is explicitly trained to distinguish
controller-specific regimes for online policy switching.
\section{Problem Definition: Challenge of Long-Horizon Switching}

\rev{We consider online switching between two fixed whole-body controllers, $\Pi=\{\pi_D,\pi_C\}$, where $\pi_D$ and $\pi_C$ denote the decoupled and coupled controllers, respectively. An episode consists
of a long-horizon motion sequence $\tau=(m_1,\ldots,m_N)$ formed by concatenating multiple motion
segments. At each high-level decision step $t$, the switching policy $\pi_{\mathrm{sw}}(c_t\mid s_t)$ selects a controller index $c_t\in\mathcal{C}=\{D,C\}$ and executes the corresponding low-level policy $\pi_{c_t}$ for $H_o$ steps. Under the options framework \cite{Sutton1999}, the resulting option value is
\begin{equation}
Q^{\pi_{\mathrm{sw}}}(s_t,c_t)
=
\mathbb{E}\left[
\sum_{i=0}^{H_o-1}\gamma^i r_{t,i}
+
\gamma^{H_o}V^{\pi_{\mathrm{sw}}}(s_{t+1})
\right],
\end{equation}
where $H_o$ is the option duration, $r_{t,i}$ is the reward at its
$i$-th low-level step, $\gamma$ is the per-low-level-step discount factor,
$s_t$ is the environment state at high-level decision $t$, and
$s_{t+1}$ is the next high-level state after executing the selected
controller for $H_o$ low-level steps.
$V^{\pi_{\mathrm{sw}}}$ denotes the continuation value under the
high-level switching policy.}

\rev{Learning this switching policy is challenging because motion demonstrations provide neither controller-selection nor switch-timing labels, while the consequences of a controller choice may be delayed
and depend on the evolving robot state. Moreover, decision-critical phases such as takeoff, landing, and recovery occur infrequently, and the number of possible motion sequences grows combinatorially with
the motion repertoire and sequence length. These challenges make reward-only switching and exhaustive training coverage impractical.}

\section{Decoupled and Coupled Whole-Body Policies}
\label{background}

\subsection{Decoupled and Coupled Whole-Body Policy}
\rev{
We adopt the decoupled whole-body reinforcement learning framework of FALCON~\cite{zhang2025falcon}, excluding its force-compensation and privileged-force modules. We leverage its decoupled upper- and lower-body training, where the upper body is conditioned on joint-space references, while
the lower body is commanded through locomotion-level quantities. The decoupled policy is denoted as $\mathbf{u}_t^D=\pi_D(\mathbf{o}_t^D)$, where $\mathbf{o}_t^D$ additionally contains proprioceptive states and short observation/action histories. The 29-DoF action is mapped to the joint target as $\mathbf{q}_t^{*,D}=\mathbf{q}^{0,D}+s_D\mathbf{u}_t^D$ and tracked using joint-level PD control.}

\rev{
We used TWIST~\cite{ze2025twist} as the
coupled whole-body controller. The coupled policy is denoted as $\mathbf{u}_t^C=\pi_C(\mathbf{o}_t^C)$, where $\mathbf{o}_t^C$ contains the whole-body motion reference, proprioceptive states, and temporal history. The 23-DoF action is mapped to the corresponding body joints of the 29-DoF G1 model
as $\mathbf{q}_{t,\mathcal{B}}^{*,C}
=\mathbf{q}_{\mathcal{B}}^{0,C}+s_C\mathbf{u}_t^C$, while the remaining hand joints are held at their default targets. The resulting whole-body target is tracked using joint-level PD control.}


    

\begin{figure}[t]
    \centering
    \includegraphics[width=\linewidth]
{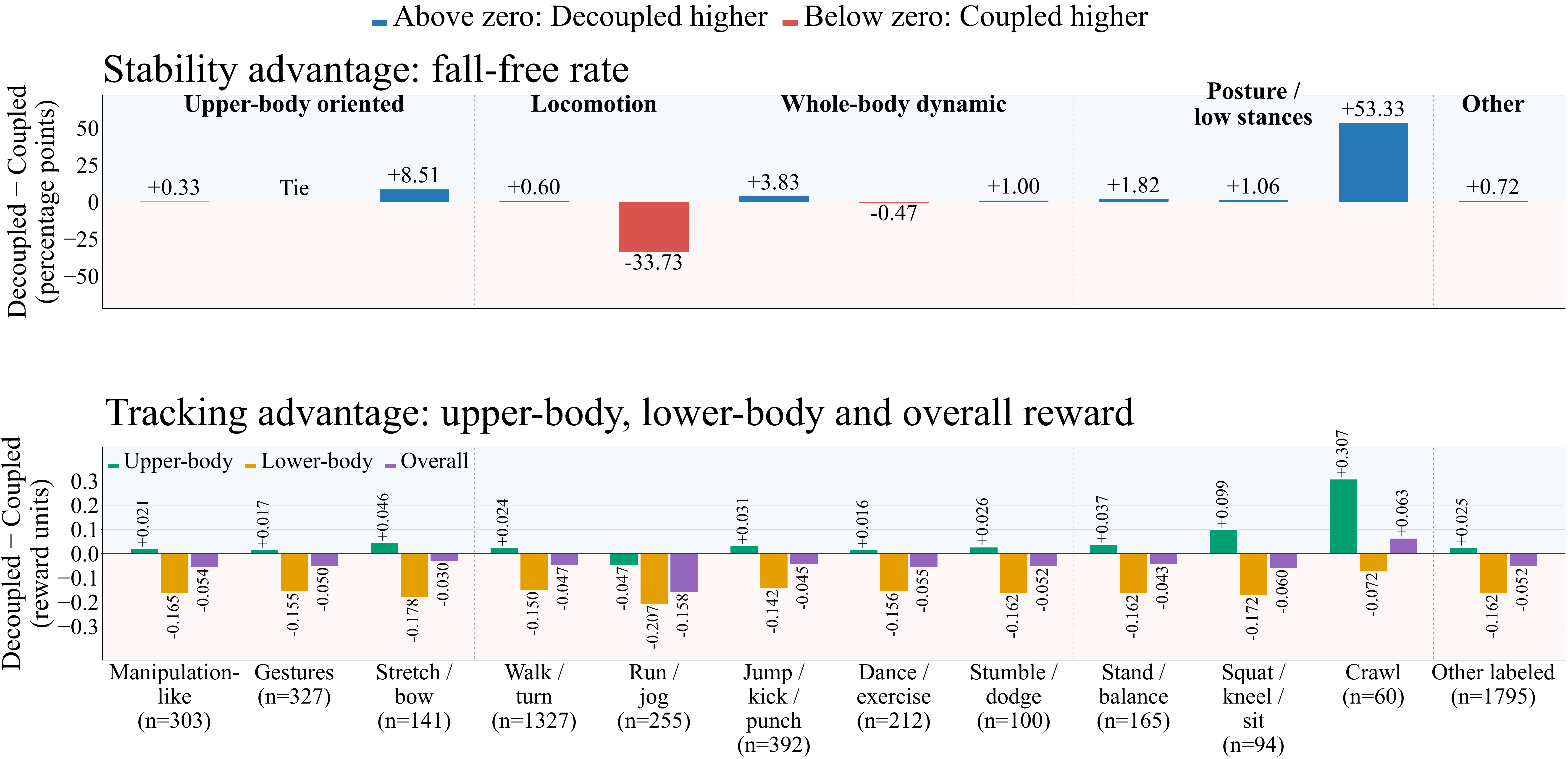}
    \caption{
    \rev{
    Complementary policy strengths across motion families.
  $\pi_D$ generally achieves higher upper-body tracking rewards,
  while $\pi_C$ achieves higher lower-body tracking rewards and
  greater running/jogging stability.
  The 5,171 motions with valid BABEL sequence-level action labels
  are grouped into 12 motion groups.
  Bars show mean $\pi_D-\pi_C$ differences in fall-free rate
  (top, percentage points) and episode-mean tracking rewards
  (bottom: green, upper-body; orange, lower-body; purple, overall).
  Positive values favor $\pi_D$; negative values favor $\pi_C$.
  $n$ denotes the number of motions.
    }}
    \label{fig:pi_DC}
    \vspace{-1em}
\end{figure}
\section{\Method{}: Balancing Agility and Stability via Online Predictive Policy Switching}
\label{method}

\begin{figure*}[t]
\vspace{0.2em}
\centering
\includegraphics[width=15.5cm]{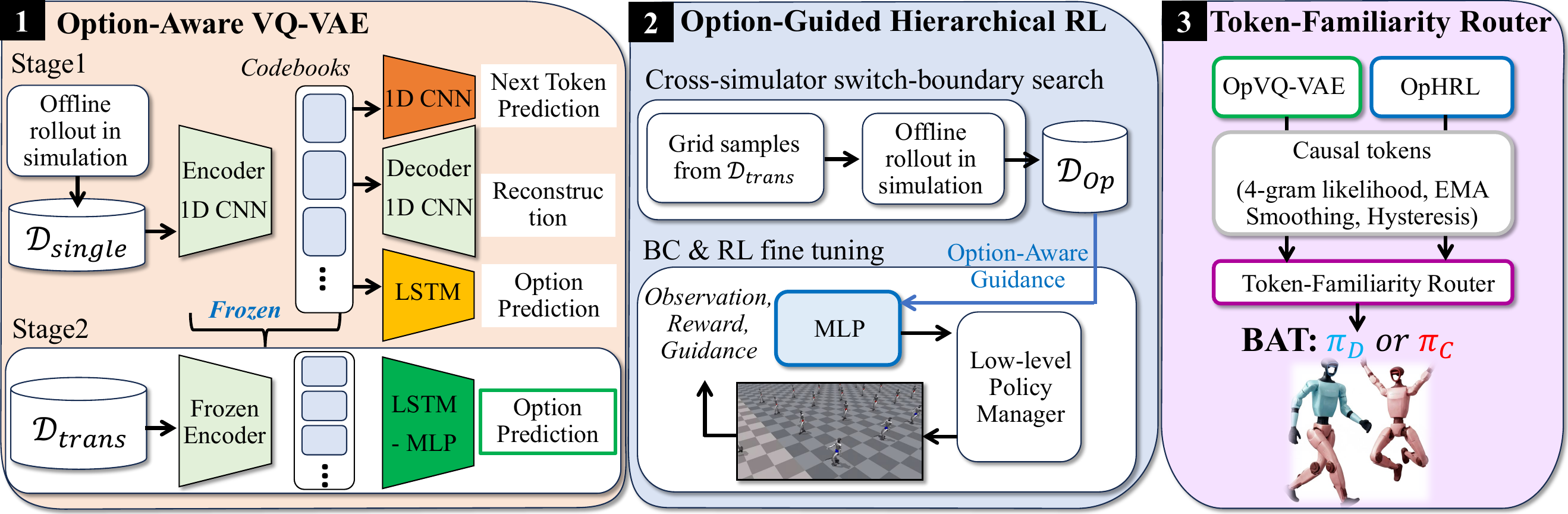}
\caption{
\rev{\textbf{Overview of BAT.}
(1) \textbf{Option-Aware VQ-VAE (OpVQ-VAE)}
learns discrete motion tokens that encode motion structure
and controller preference through reconstruction,
option-prediction, and auxiliary objectives.
After freezing the encoder and codebook, an LSTM--MLP head
learns transition-aware option prediction from individual
and composed motions.
(2) \textbf{Option-Guided Hierarchical RL (OpHRL)}
uses cross-simulator switch-boundary supervision from
$\mathcal{D}_{\mathrm{Op}}$ for behavior-cloning initialization
and guided, proprioception-aware RL refinement.
The supervision is obtained through full-chain Isaac Gym
rollouts and MuJoCo validation
(Sec.~\ref{sec:data_supervision}).
A low-level policy manager executes the selected controller.
(3) \textbf{Token-Familiarity Router (TFR)}
estimates causal token-sequence familiarity using 4-gram
likelihood and selects between the predictions of
OpVQ-VAE and OpHRL.
EMA smoothing and hysteresis are applied to the familiarity
score and routing decision, yielding the final selection
between the decoupled policy $\pi_D$ and coupled policy $\pi_C$.}}
\vspace{-1em}
\label{bat_fig_flow}
\end{figure*}

We first analyze the complementary strengths of $\pi_D$ and
$\pi_C$ across motion families (Fig.~2): $\pi_D$ generally
provides stronger upper-body tracking, whereas $\pi_C$ provides
stronger lower-body tracking and greater running/jogging stability.
We then introduce BAT, which combines these strengths through
online policy switching (Fig.~3).

\subsection{Training Data and Cross-Simulator Switching Supervision}
\label{sec:data_supervision}

\rev{We construct $\mathcal{D}_{\mathrm{single}}$ by filtering AMASS
for physically feasible motions and split it at the motion level
into a 90\% training subset,
$\mathcal{D}_{\mathrm{single}}^{\mathrm{train}}$,
and a 10\% held-out subset,
$\mathcal{D}_{\mathrm{single}}^{\mathrm{test}}$.
Each frame receives a preferred-controller label
$c_t^\star\in\mathcal{C}$ by evaluating the two frozen policies
from the same robot state in MuJoCo.
The training subset supports Stage-1 option-aware tokenizer learning.}

\rev{We construct $\mathcal{D}_{\mathrm{trans}}$ by randomly pairing
and concatenating training clips.
Each chain $\tau_i=(m_i^A,m_i^B)$ has an original clip boundary
$b_i^{\mathrm{clip}}$ and a controller order $(c_i^A,c_i^B)$
assigned through offline motion evaluation.
For OpVQ-VAE Stage~2, each clip retains its controller label,
and the option head is trained on
$\mathcal{D}_{\mathrm{trans}}
\cup\mathcal{D}_{\mathrm{single}}^{\mathrm{train}}$
with the tokenizer frozen.
Tokens from $\mathcal{D}_{\mathrm{trans}}$ also provide the
training statistics for familiarity estimation and threshold
calibration in TFR; no test data are used.}

\rev{To obtain switching supervision for OpHRL, we refine the
clip boundaries through cross-simulator search while preserving
the controller order.
For each chain, we evaluate 21 candidate boundaries
$b_{i,\delta}=b_i^{\mathrm{clip}}+\delta$, where
$\delta=g+a$,
$g\in\{-60,-30,-15,0,15,30,60\}$, and
$a\in\{-5,0,5\}$, measured in motion frames at 30~Hz.
All candidates undergo full-chain Isaac Gym rollouts.
MuJoCo then reevaluates $\delta=0$ and up to four successful
shifted candidates.
Among MuJoCo-successful candidates, selection prioritizes
Isaac Gym success, proximity to the clip boundary,
MuJoCo and Isaac Gym rewards, and tracking performance.
If none succeeds in MuJoCo, we retain the original boundary.}

\rev{The selected boundary $b_i^\star$ defines controller labels
$\mathbf{c}_i^\star$, assigning $c_i^A$ before the boundary
and $c_i^B$ thereafter.
The resulting dataset,
$\mathcal{D}_{\mathrm{Op}}
=\{(\tau_i,\mathbf{c}_i^\star,b_i^\star)\}_{i=1}^{M}$,
provides supervision for OpHRL's behavior-cloning initialization
and subsequent closed-loop RL refinement.}

\subsection{Option-Aware Vector-Quantized VAE}
\label{sec:vqvae_switching}
\rev{The number of possible motion sequences grows combinatorially as individual motions are concatenated.
Inspired by discrete motion vocabularies and tokenized motion embeddings~\cite{zhang2023generating,jiang2023motiongpt,kim2025moreflow}, we adopt a shared token vocabulary to represent diverse motions and their sequential compositions through reusable tokens, reducing the need to cover every motion combination during training.
However, prior discrete motion representations primarily target reconstruction, generation, or prediction~\cite{oord2017neural,zhang2023generating,guo2024momask,radosavovic2024humanoid}, without explicitly encoding which low-level controller is more suitable.
We therefore propose an option-aware VQ-VAE (OpVQ-VAE) that learns reusable motion tokens encoding both motion structure and controller preference.
To make the tokens informative for controller selection, we augment motion reconstruction with option-prediction supervision and latent separation. Using the datasets defined in Sec.~\ref{sec:data_supervision},
we train OpVQ-VAE in two stages.
Stage~1 learns option-aware tokens from individual motions,
while Stage~2 freezes the tokenizer and trains a
transition-aware option head on individual and composed motions.}

\subsubsection{Motion Tokenization}

\rev{
Each frame is represented by a 50-dimensional feature vector $x_t\in\mathbb R^{50}$, consisting of root-local translational deltas, a yaw delta, 23-DOF joint positions, and 23-DOF joint velocities. The root deltas are obtained from smoothed root velocities expressed in the local root frame. From each clip in $\mathcal{D}_{\mathrm{single}}^{\mathrm{train}}$,
we extract causal windows
$W_t=(x_{t-W+1},\ldots,x_t)$ of length $W=16$ and normalize them
element-wise using the training-set statistics $\mu$ and $\sigma$.
The encoder and VQ layer map each window to a discrete token sequence $\mathbf z_t=(z_{t,1},\ldots,z_{t,L})$, where
$z_{t,\ell}\in\{1,\ldots,K\}$, by nearest-neighbor assignment in the codebook
$\mathbf E\in\mathbb R^{K\times d_z}$.
We use $K=d_z=256$. With residual convolutional blocks and $4\times$ temporal downsampling, each $W=16$ window produces $L=4$ tokens.
The decoder reconstructs the window as
$\hat{\mathcal W}_t=\mathrm{Dec}(\mathbf e_{\mathbf z_t})$, where $\mathbf e_{\mathbf z_t} =(\mathbf E_{z_{t,1}},\ldots,\mathbf E_{z_{t,L}})$ denotes the selected codebook embeddings.
}

\subsubsection{Stage 1: Option-Aware Tokenizer Training}
\label{sec:base_tokenizer}

\rev{
We augment an EMA-based VQ-VAE with option-prediction, latent-separation, and
token-prediction objectives. Let $\mathbf z_e$ denote the continuous encoder
output. The base objective is
}
\begin{equation}
\mathcal L_{\mathrm{VQ}}
=
\mathcal L_{\mathrm{rec}}
+\alpha_c\mathcal L_{\mathrm{commit}}
+\alpha_v\mathcal L_{\mathrm{vel}},
\label{eq:vq_loss}
\end{equation}
\rev{
where
$\mathcal L_{\mathrm{rec}}
=\|\mathcal W_t-\hat{\mathcal W}_t\|_H$,
$\mathcal L_{\mathrm{commit}}
=\|\mathbf z_e-\mathrm{sg}[\mathbf e_{\mathbf z_t}]\|_2^2$, and
$\mathcal L_{\mathrm{vel}}
=\|\Delta\mathcal W_t-\Delta\hat{\mathcal W}_t\|_H$.
Here, $\|\cdot\|_H$ denotes the Smooth-$\ell_1$ loss, $\Delta$ denotes
frame-to-frame differences, and $\mathrm{sg}$ denotes stop-gradient.
The codebook is updated by exponential moving average.
The auxiliary heads use
$\tilde{e}_{z_t}=z_e+\mathrm{sg}[e_{z_t}-z_e]$,
the straight-through quantized embeddings.
We set $(\alpha_c,\alpha_v)=(0.02,0.5)$.}

\rev{
\noindent\textbf{Option prediction.}
Reconstruction alone does not ensure that the motion tokens retain information relevant to option selection. We therefore use an LSTM--MLP head to predict the preferred-option distribution
$p_t^{\mathrm{S1}}(c)
= p_\phi(c \mid \tilde{e}_{z_t})$
from the complete token sequence of the causal $W$-frame window ending at
time $t$. We assign each window the majority frame-level controller label
$\bar{c}_t^\star=\arg\max_{c\in\mathcal{C}}
\frac{1}{W}\sum_{s=t-W+1}^{t}\mathbf{1}[c_s^\star=c]$
and optimize
$\mathcal{L}_{\mathrm{pol}}
= \operatorname{CE}
\left(p_t^{\mathrm{S1}}(\cdot), \bar{c}_t^\star\right)$. This objective makes the causal token sequence predictive of the preferred option.}

\rev{
\noindent\textbf{Auxiliary latent objectives.}
We use latent separation and next-token prediction
to encourage option-discriminative tokens and capture
local motion progression. For each window, we average its straight-through token
embeddings and group the resulting representations by
their majority controller labels.
Let $\bar{e}_D$ and $\bar{e}_C$ denote the corresponding
minibatch centroids. When both classes are present, we minimize $\mathcal{L}_{\mathrm{sep}}
=\cos(\bar{e}_D,\bar{e}_C)$. The next-token loss is
$\mathcal{L}_{\mathrm{lat}}
=\mathrm{CE}(g_\psi(\tilde{e}_{t,1:L-1}),z_{t,L})$,
where $g_\psi$ predicts the final token from the
preceding straight-through embeddings. The complete Stage-1 objective is
\begin{equation}
\mathcal{L}_{\mathrm{OpVQ}}
=
\mathcal{L}_{\mathrm{VQ}}
+\alpha_p\mathcal{L}_{\mathrm{pol}}
+\alpha_s\mathcal{L}_{\mathrm{sep}}
+\alpha_l\mathcal{L}_{\mathrm{lat}},
\end{equation}
where $(\alpha_p,\alpha_s,\alpha_l)=(0.5,0.05,0.1)$.}

\subsubsection{Stage 2: Transition-Aware Option Head Training}
\label{sec:transition_head}

\rev{With the tokenizer frozen, we initialize the Stage-2 option head from Stage~1 and train it on $\mathcal{D}_{\mathrm{trans}}
\cup\mathcal{D}_{\mathrm{single}}^{\mathrm{train}}$. Given causal token embeddings $e_{z_t}$, the head predicts $p_t^{S2}(c)=p_\theta(c\mid e_{z_t})$ and selects
$\hat{c}_t=\arg\max_{c\in\mathcal{C}}p_t^{S2}(c)$. The Stage-2 training objective is:}
\rev{\begin{equation}
\mathcal{L}_{\mathrm{trans}}
=
\lambda_p\operatorname{CE}
\left(p_t^{\mathrm{S2}},c_t^{\mathrm{S2}}\right)
+\lambda_c\mathcal{L}_{\mathrm{cons}}
+\lambda_s\mathcal{L}_{\mathrm{sw}}.
\label{eq:transition_loss}
\end{equation}}
\rev{
Here, $c_t^{S2}$ is the controller label of the constituent
motion clip containing frame $t$, as defined in
Sec.~\ref{sec:data_supervision}.
We use $\mathcal{L}_{\mathrm{cons}}
=\|p_t^{\mathrm{S2}}-p_{t-1}^{\mathrm{S2}}\|_2^2$
to encourage temporal consistency and
$\mathcal{L}_{\mathrm{sw}}
=p_t^{\mathrm{S2}}(D)p_{t-1}^{\mathrm{S2}}(C)
+p_t^{\mathrm{S2}}(C)p_{t-1}^{\mathrm{S2}}(D)$
to discourage frequent controller changes.
We set $(\lambda_p,\lambda_c,\lambda_s)=(1,0.1,0.1)$.}

\subsection{Option-Guided Hierarchical Reinforcement Learning}
\label{sec:ophrl}
\rev{OpHRL uses the cross-simulator supervision in
$\mathcal{D}_{\mathrm{Op}}$ to initialize and guide
proprioception-aware, closed-loop switching.
Unlike OpVQ-VAE, which uses only motion references,
OpHRL observes both the robot state and motion reference
to select between the frozen policies $\pi_D$ and $\pi_C$.}

\subsubsection{Observation and Action Space}
The high-level policy $\pi_{\mathrm{OpHRL}}$ receives a stacked observation constructed from proprioception, motion reference, and policy state. A single-step observation is
$\bar{o}_t =
[v_t,\omega_t,g_t,q_t^{\mathrm{ref}},
q_t,\dot{q}_t,\mathbf{c}_{t-1},a_{t-1},\eta_t]$, where $v_t,\omega_t,g_t\in\mathbb R^3$ are the base
linear and angular velocities and projected gravity;
$q_t^{\mathrm{ref}},q_t,\dot q_t,a_{t-1}\in\mathbb R^{29}$ are the reference joint target, joint position, joint velocity, and previous low-level action;
$\mathbf{c}_{t-1}\in\{0,1\}^2$
is the one-hot encoding of the previous controller selection; and $\eta_t \in \mathbb{R}^3$ is the current policy state including transition phase. We stack five consecutive steps to form $o_t = [\bar{o}_t, \dots, \bar{o}_{t-4}] \in \mathbb{R}^{650}$. The action space is discrete, $\mathcal{A}=\{0,1\}$, corresponding to selecting $\pi_D$ or $\pi_C$, with $\pi_{\mathrm{OpHRL}}$ outputting logits $\ell_t \in \mathbb{R}^2$ parameterizing the categorical distribution $\pi_{\mathrm{OpHRL}}(c_t \mid o_t)$.

\subsubsection{Reward and Optimization}

The high-level policy $\pi_{\mathrm{OpHRL}}$ is trained with the composite reward
$r_t=w_{\mathrm{track}}r_{\mathrm{track}}
-w_{\mathrm{reg}}p_{\mathrm{reg}}
-w_{\mathrm{sw}}p_{\mathrm{sw}}
-w_{\mathrm{term}}p_{\mathrm{term}}$. The tracking reward follows the deploy-time evaluation metric,
$r_{\mathrm{track}}
= w_u r_{\mathrm{upper}}
+ w_l r_{\mathrm{lower}}
+ w_v r_{\mathrm{vel}}
+ w_h r_{\mathrm{height}}$.
For the upper- and lower-body joint groups, indexed by $j\in\{u,l\}$, we use
$r_j=\exp[-\alpha_j\operatorname{MSE}
(\mathbf q_t^j,\mathbf q_t^{\mathrm{ref},j})]$.
The base-velocity and root-height rewards are
\begin{equation}
\begin{aligned}
r_{\mathrm{vel}}
={}&
\frac{1}{2}
\sum_{i\in\{x,y\}}
\exp\!\left[
-\frac{
(v_{i,t}^{\mathrm{cmd}}-v_{i,t})^2
}{
\sigma_v
}
\right],\\
r_{\mathrm{height}}
={}&
\exp\!\left[
-\alpha_h
(z_t-z_t^{\mathrm{ref}})^2
\right].
\end{aligned}
\label{eq:base_tracking_reward}
\end{equation}

The regularization penalty is
\begin{equation}
\begin{aligned}
p_{\mathrm{reg}}
={}&
\beta_\tau\|\tau_t\|_2^2
+\beta_{\dot q}\|\dot{\mathbf q}_t\|_2^2\\
&+
\beta_{\ddot q}\|\ddot{\mathbf q}_t\|_2^2
+\beta_\Delta\|a_t-a_{t-1}\|_2^2\\
&+
\beta_\omega\|\boldsymbol\omega_t^{xy}\|_2^2
+\beta_f p_{\mathrm{foot}}
+\beta_d p_{\mathrm{drift}}.
\end{aligned}
\label{eq:regularization_penalty}
\end{equation}

where $p_{\mathrm{foot}}$ aggregates contact-force, foot-slip, and stumble penalties, and $p_{\mathrm{drift}}$ penalizes linear and angular drift under near-stationary commands. The switching penalty $p_{\mathrm{sw}}$ penalizes each
controller change once the cumulative switch count
exceeds the number of clip boundaries in the motion chain.
The termination penalty $p_{\mathrm{term}}$ adds a unit
penalty for a fall and another for terminal instability,
defined as low base height or excessive body tilt
near the end of the motion.

We initialize $\pi_{\mathrm{OpHRL}}$ by behavior cloning
with labels from $\mathcal{D}_{\mathrm{Op}}$ and retain
the BC loss during PPO fine-tuning:
\begin{equation}
\mathcal{L}_{\mathrm{BC}}
=
-\mathbb{E}_t
\left[
\log \pi_{\mathrm{OpHRL}}(c_t^\star \mid o_t)
\right].
\end{equation}
Here, $o_t$ is the current high-level observation, and
$c_t^\star=c_{i,f_t}^\star$ is the label retrieved from
$\mathcal{D}_{\mathrm{Op}}$ for motion chain $i$
at its current motion-frame index $f_t$. The complete objective is
\begin{equation}
\begin{aligned}
\mathcal{L} ={}&
\mathcal{L}_{\mathrm{PPO}}
+\lambda_{\mathrm{BC}}(k)\mathcal{L}_{\mathrm{BC}} \\
&+\lambda_{\mathrm{KL}}\,
\mathbb{E}_t\!\left[
D_{\mathrm{KL}}\!\left(
\pi_{\mathrm{OpHRL}}(\cdot\mid o_t)
\,\Vert\,
\pi_{\mathrm{BC}}(\cdot\mid o_t)
\right)
\right].
\end{aligned}
\label{eq:hrl_objective}
\end{equation}
Here, $\pi_{\mathrm{BC}}$ is the frozen policy used
to initialize PPO fine-tuning.
The coefficient $\lambda_{\mathrm{BC}}(k)$ decreases
with training iteration $k$, gradually reducing
the contribution of BC supervision.

\subsection{Token-Familiarity Router} 
\rev{Unlike conventional gates that directly predict or blend controller
outputs, our TFR resolves disagreements between OpHRL and OpVQ-VAE
using token-sequence familiarity. OpHRL captures execution-state feedback
from composed sequences in $\mathcal D_{\mathrm{Op}}$, but cannot cover
the combinatorial composition space, whereas OpVQ-VAE generalizes from
causal motion tokens without observing the robot state.} 

\rev{To select between
these complementary predictors, we estimate familiarity using a lightweight causal $m$-gram
model fitted to tokens from $\mathcal{D}_{\mathrm{trans}}$
(Sec.~\ref{sec:data_supervision}), following likelihood-based
assessment of discrete representations~\cite{oord2017neural,marimont2021anomaly}. At each motion-reference frame $f$, the causal motion
window produces a token row
$z_f=(z_{f,1},\ldots,z_{f,L})$.
We concatenate the most recent 32 rows in temporal order as
$z_f^{\mathrm{flat}}
=(z_{f-31},\ldots,z_f)
=(\zeta_1,\ldots,\zeta_{N_{\mathrm{tok}}})$,
where $N_{\mathrm{tok}}=32L$.
The router defaults to OpHRL until all 32 rows are available. Let $\mathcal{G}_f$ contain the token positions with
complete causal $m$-gram contexts.
Conditional probabilities use additive smoothing,
$P_{\mathrm{train}}(z\mid h)
=(C(h,z)+0.1)/(C(h)+0.1K)$,
where $h$ is the preceding three-token context
and $C$ denotes training-token counts.
We define the token negative log-likelihood score as}
\rev{\begin{equation}
u_f =
-\frac{1}{|\mathcal{G}_f|}
\sum_{j\in\mathcal{G}_f}
\log P_{\mathrm{train}}
\left(\zeta_j\mid\zeta_{j-m+1:j-1}\right).
\end{equation}
We use $m=4$.
Lower $u_f$ indicates greater token-sequence familiarity.
The raw window-NLL is smoothed as
$\bar{u}_f=\beta u_f+(1-\beta)\bar{u}_{f-1}$,
with $\beta=0.05$.
At each high-level decision, the router uses
$\bar{u}_{f_t}$, where $f_t$ is the current motion-frame
index.
The hysteretic router $g_t\in\{0,1\}$ selects OpHRL
when $g_t=1$ and OpVQ-VAE when $g_t=0$:
}
\begin{equation}
g_t =
\begin{cases}
0, & g_{t-1}=1 \land \bar{u}_{f_t}>\tau_{\mathrm{off}},\\
1, & g_{t-1}=0 \land \bar{u}_{f_t}\leq\tau_{\mathrm{on}},\\
g_{t-1}, & \mathrm{otherwise},
\end{cases}
\quad
\tau_{\mathrm{on}}<\tau_{\mathrm{off}}.
\label{eq:bat_gate}
\end{equation}
\rev{
We set $\tau_{\mathrm{on}}=1.75$ and
$\tau_{\mathrm{off}}=2.69$ to the 95th and 99.5th percentiles
of the training EMA-smoothed window-NLL distribution,
respectively.}

\begin{figure}[t]
    \centering
    \begin{minipage}[t]{\columnwidth}
        \centering
        \includegraphics[width=0.8\linewidth]{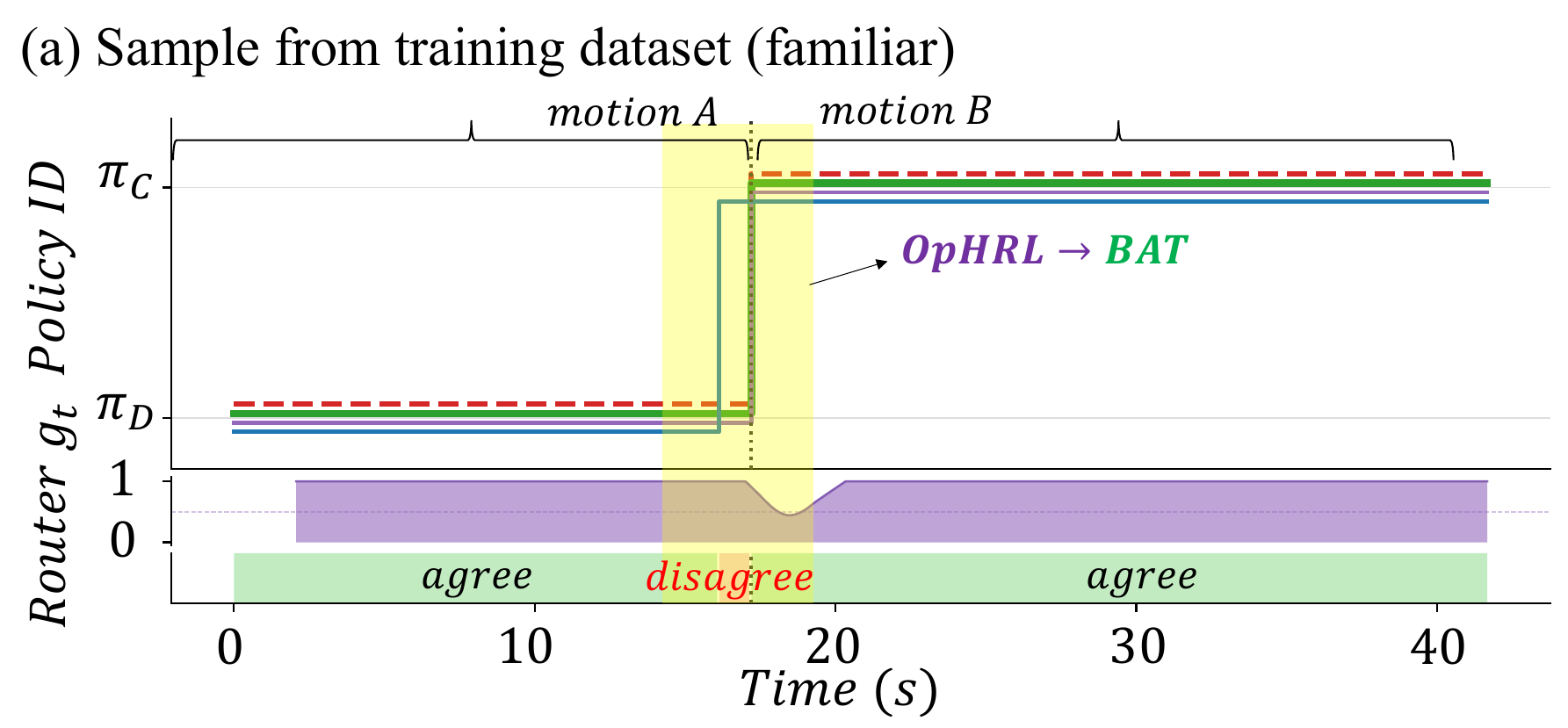}
        \vspace{-1mm}
    \end{minipage}
    \begin{minipage}[t]{\columnwidth}
        \centering
        \includegraphics[width=0.8\linewidth]{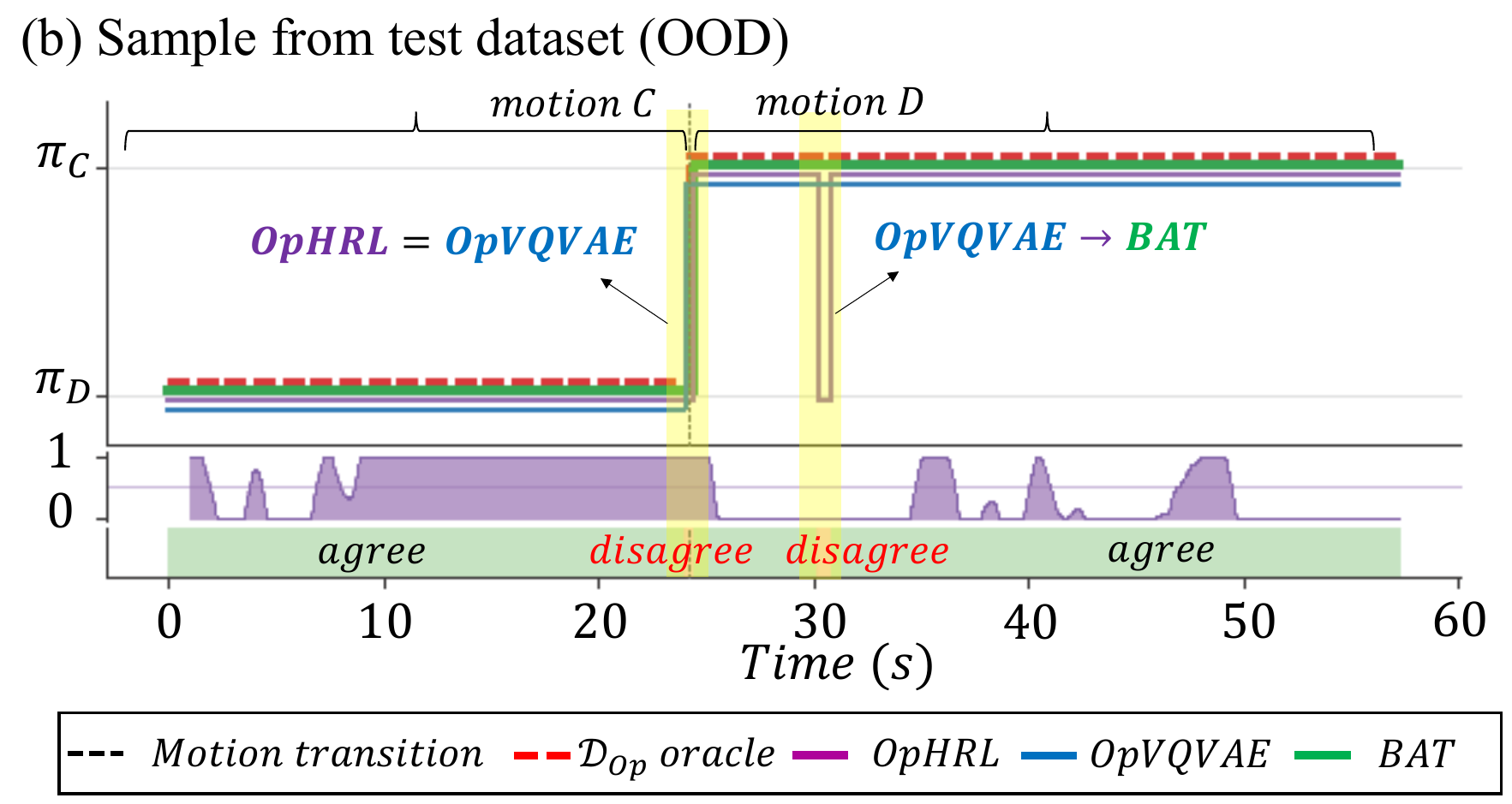}
        \vspace{-1mm}
        \footnotesize (a)
    \end{minipage}

    \vspace{1mm}

    \begin{minipage}[t]{0.48\columnwidth}
        \centering
        \includegraphics[width=\linewidth]
        {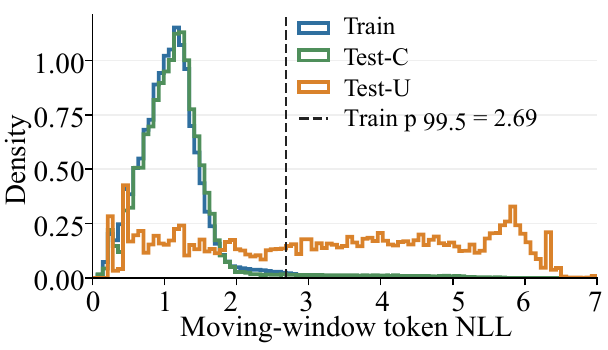}
        \vspace{-1mm}
        \footnotesize (b)
    \end{minipage}%
    \hfill
    \begin{minipage}[t]{0.48\columnwidth}
        \centering
        \includegraphics[width=\linewidth]
        {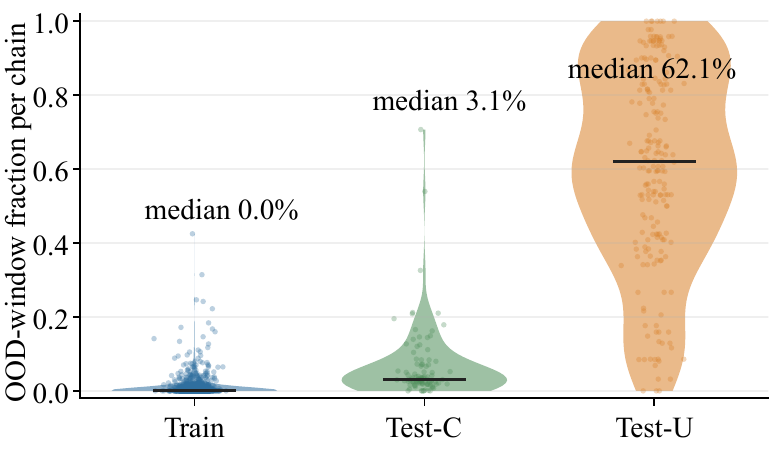}
        \vspace{-1mm}
        \footnotesize (c)
    \end{minipage}

    \caption{
    \rev{\textbf{Token familiarity resolves predictor disagreements by
    selecting OpHRL for familiar sequences and OpVQ-VAE for
    unfamiliar ones.}
    (a) Switching traces for a training chain (top) and a Test-U
    chain containing unseen motions (bottom). Highlighted regions
    show disagreements: \Method{} follows OpHRL when $g_t=1$
    (familiar) and OpVQ-VAE when $g_t=0$ (unfamiliar).
    When both predictors agree, their common prediction is used.
    Dashed vertical lines mark motion-clip boundaries.
    (b) Moving-window token-NLL distributions for Train, Test-C
    (seen motions in unseen combinations), and Test-U
    (unseen motions in unseen combinations).
    The hysteresis thresholds $\tau_{\mathrm{on}} = 1.75$ and $\tau_{\mathrm{off}} = 2.69$ correspond to the 95th and 99.5th percentiles of the training EMA-smoothed window-NLL distribution.
    (c) Per-chain fractions of unfamiliar windows, with medians
    of 0.0\%, 3.1\%, and 62.1\%, respectively, indicating that
    the router responds primarily to unseen motions.
    }}
    \label{fig:router_analysis}
    \vspace{-0.5cm}
\end{figure}

\rev{Let $c_t^{\mathrm{OpHRL}}$ and $c_t^{\mathrm{OpVQ}}$ denote the option IDs
proposed by OpHRL and OpVQ-VAE. Unlike prior likelihood-based anomaly detection, BAT uses token familiarity only as a reliability cue when the two predictors disagree:
\begin{equation}
c_t^{\mathrm{BAT}} =
\begin{cases}
c_t^{\mathrm{OpHRL}},
& c_t^{\mathrm{OpHRL}} = c_t^{\mathrm{OpVQ}},\\
c_t^{\mathrm{OpHRL}},
& c_t^{\mathrm{OpHRL}} \neq c_t^{\mathrm{OpVQ}} \land g_t=1,\\
c_t^{\mathrm{OpVQ}},
& c_t^{\mathrm{OpHRL}} \neq c_t^{\mathrm{OpVQ}} \land g_t=0.
\end{cases}
\label{eq:bat_fusion}
\end{equation}
The router selects OpHRL for familiar sequences and OpVQ-VAE for unfamiliar ones (Fig.~\ref{fig:router_analysis}(a)).
The median OOD-window rate rises from $0.0\%$ on Train and $3.1\%$ on
Test-C to $62.1\%$ on Test-U (Fig.~\ref{fig:router_analysis}(b,c)),
showing that the router primarily detects unseen motions rather than
unseen combinations. This familiarity-aware routing outperforms either
predictor alone.}

\section{Experiment}
\label{Experiment}

To evaluate \Method{}, we conduct extensive benchmark validation in simulation and demonstrate real-world deployment on hardware. We address the following key questions:
\rev{
(Q1) Does the proposed \Method{} identify higher-quality switching intervals and improve task performance relative to baselines?
(Q2) How do option-aware representation learning and discrete quantization affect generalization to unseen motion combinations and unseen motions?
(Q3) Does BAT achieve higher task success rates and more precise control than other whole-body control methods?
(Q4) Does \Method{} generalize reliably across diverse scenarios when deployed on real hardware?}

\subsection{Implementation Details and Experimental Setup}

\subsubsection{Training Details}
We train the hierarchical RL switching policy using Discrete PPO within Isaac Gym via HumanoidVerse~\cite{HumanoidVerse}, implemented as a lightweight MLP. Training takes approximately 12--15 hours on an NVIDIA L40S GPU, initialized with behavioral cloning supervision that gradually decays toward RL to enable fast early convergence. Training datasets and switching supervision are constructed
as described in Sec.~\ref{sec:data_supervision}.

\subsubsection{Simulation and Hardware Setup}
Both IsaacGym~\cite{makoviychuk2021isaac} and MuJoCo~\cite{todorov2012mujoco} are utilized for offline policy evaluation, data construction, and evaluation process. The control frequency is 50 Hz with 200 Hz physics, and policy switching is performed at 5 Hz. \rev{On an Intel Core Ultra 9 285K CPU, the complete high-level decision pipeline takes 8.493 ms on average.} For real-world validation, we deploy the policy on a Unitree G1 humanoid
without additional tuning. Standard domain randomization is applied during
training for sim-to-real transfer. The option-aware tokenizer operates on the
30~Hz motion-reference timeline. Each tokenization input is a causal window of
$W=16$ motion frames, corresponding to $0.53$~s. With $4\times$ temporal
downsampling, each window produces four VQ tokens. We generate one causal token
row at each motion-reference frame, i.e., at 30~Hz. The raw familiarity score averages 4-gram token NLLs
over the latest 32 causal token rows
($1.07\,\mathrm{s}$ at $30\,\mathrm{Hz}$)
and is then EMA-smoothed with $\beta=0.05$ \rev{(we empirically observed that controller decisions were sensitive
to 1- and 2-gram models but unchanged under the other tested settings).}

\subsubsection{Policy Switching Mechanism} 
\rev{Switching between independently trained controllers can introduce out-of-distribution histories and discontinuous targets, potentially destabilizing the robot. At each handoff, we
preserve the robot state $(\mathbf{q}_t,\dot{\mathbf{q}}_t)$, reset the
incoming policy's histories and reference states, and warm-start its previous action from the current configuration.}
\rev{
For the incoming target $\mathbf{q}_t^{+}$, we initialize $\mathbf{e}_q(0)=\mathbf{q}_t-\mathbf{q}_t^{+}$ and decay it using
$\ddot{\mathbf{e}}_q+d\dot{\mathbf{e}}_q+k\mathbf{e}_q=0$, with $d=2\sqrt{k}$. The corrected target
$\tilde{\mathbf q}_t=\mathbf q_t^{\mathrm{raw}}+\mathbf e_q(t)$ thereby
smoothly converges from the current configuration to the incoming target. The same procedure is applied to velocity targets, while the policy-specific PD gains are smoothly blended during the transition.
}

\begin{table*}[t]
\centering
\begin{threeparttable}
\small

\caption{
\rev{
Performance comparison across three evaluation splits:
\textbf{Train} (650 chains) uses seen motions in seen combinations,
\textbf{Test-C} (70 chains) uses seen motions in unseen combinations,
and \textbf{Test-U} (160 chains) uses unseen motions in unseen combinations.
Mean Switches is reported as a diagnostic of switching frequency.
Best and second-best distinct Success and Reward values across all methods, including offline oracles, are indicated by
\textbf{bold} and \underline{underlined} values, respectively;
ties receive the same formatting.
Offline oracles replay precomputed controller selections and do not guarantee successful execution.
}}
\label{tab:performance_all_splits}

\renewcommand{\arraystretch}{0.95}
\setlength{\tabcolsep}{8pt}

\begin{tabular}{
    ll
    ccc@{\hspace{4pt}}
    ccc@{\hspace{4pt}}
    ccc
}
\toprule
\multicolumn{2}{c}{}
& \multicolumn{3}{c}{\textbf{Success (\%) $\uparrow$}}
& \multicolumn{3}{c}{\textbf{Reward $\uparrow$}}
& \multicolumn{3}{c}{\textbf{Mean Switches}} \\
\cmidrule(lr){3-5}
\cmidrule(lr){6-8}
\cmidrule(lr){9-11}
\textbf{Category}
& \textbf{Method}
& \textbf{Train}
& \textbf{Test-C}
& \textbf{Test-U}
& \textbf{Train}
& \textbf{Test-C}
& \textbf{Test-U}
& \textbf{Train}
& \textbf{Test-C}
& \textbf{Test-U} \\
\midrule

Ours
& \color{green!50!black} BAT
& \color{green!50!black} \underline{85.2}
& \color{green!50!black} \textbf{70.0}
& \color{green!50!black} \textbf{71.3}
& \color{green!50!black} 127.9
& \color{green!50!black} 47.10
& \color{green!50!black} \textbf{115.5}
& \color{green!50!black} 1.43
& \color{green!50!black} 1.93
& \color{green!50!black} 2.16 \\
\midrule

\multirow{6}{*}{Ablations}
& OpVQ-VAE
& 77.8 & 65.7 & \underline{70.6}
& 124.0 & \textbf{58.88} & \textbf{115.5}
& 1.04 & 1.01 & 1.03 \\

& OpHRL
& 80.8 & \underline{68.6} & 41.3
& 117.0 & 47.17 & 66.1
& 1.54 & 1.83 & 3.61 \\

& BC
& 71.8 & 62.9 & 40.6
& 107.0 & 57.34 & 66.1
& 2.39 & 2.21 & 3.84 \\

& Continuous Op-AE
& 78.8 & 67.1 & 61.9
& 51.65 & 45.10 & 45.09
& 0.88 & 0.80 & 1.40 \\

& VQ-VAE
& 69.8 & 65.7 & 49.4
& 47.60 & 45.66 & 39.81
& 1.34 & 1.19 & 1.61 \\

& Vanilla HRL
& 0.0 & 0.0 & 0.0
& -2.5 & -2.55 & -2.2
& 6.61 & 5.98 & 7.15 \\
\midrule

\multirow{4}{*}{Baselines}
& MoE
& 66.8 & 57.1 & 58.8
& 67.5 & 43.73 & 58.8
& 0.37 & 0.29 & 0.36 \\

& Learned Gate
& 66.8 & 47.1 & 36.9
& 107.7 & 38.10 & 60.5
& 0.37 & 0.10 & 0.12 \\

& Causal TF
& 50.2 & 45.7 & 39.4
& 79.9 & 36.61 & 64.8
& 0.21 & 0.19 & 0.06 \\

& Option-Critic
& 48.8 & 45.7 & 30.6
& 39.1 & 35.86 & 33.8
& 0.00 & 0.00 & 0.00 \\
\midrule

\multirow{2}{*}{Fixed Policies}
& Decoupled
& 52.2 & 50.0 & 30.0
& 84.0 & 38.82 & 48.6
& 0.00 & 0.00 & 0.00 \\

& Coupled
& 29.2 & 28.6 & 23.8
& 47.7 & 24.50 & 39.9
& 0.00 & 0.00 & 0.00 \\
\midrule

\multirow{2}{*}{Offline Oracles}
& $\mathcal{D}_{\mathrm{Op}}$ oracle
& \textbf{86.5} & -- & --
& \textbf{151.6} & -- & --
& 0.89 & -- & -- \\

& $\mathcal{D}_{\mathrm{trans}}$ oracle
& 82.8 & \underline{68.6} & 69.4
& \underline{143.2} & \underline{57.74} & \underline{112.5}
& 0.89 & 0.86 & 0.94 \\

\bottomrule
\end{tabular}
\end{threeparttable}
\end{table*}
\begin{table*}[t]
\centering

\begin{threeparttable}

\caption{\rev{
Comparison of motion tracking performance and perturbation robustness across
diverse motions. \Method{} consistently selects the decoupled policy $\pi_D$
for the standing and static motions used in the perturbation tests.
}}
\label{tab:method_comparison}

\small
\begin{tabular}{lccccc}
\toprule
\textbf{Metric}
& \textbf{GMT \cite{chen2025gmt}}
& \textbf{FALCON \cite{zhang2025falcon}}
& \textbf{TWIST \cite{ze2025twist}}
& \textbf{SONIC \cite{luo2025sonic}}
& \textbf{BAT (Ours)} \\
\midrule

Success ($n$ / total)
& 7149 / 8261
& \underline{7735 / 8261}
& 7659 / 8261
& 7519 / 8261
& \textbf{7901 / 8261} \\

Success Rate (\%)
& 86.54
& \underline{93.63}
& 92.71
& 91.02
& \textbf{95.64} \\

\midrule

Mean Reward
& 0.6347
& 0.7677
& \underline{0.8002}
& 0.7343
& \textbf{0.8468} \\

Upper-body tracking
& 0.8695
& \textbf{0.9776}
& 0.9483
& 0.8858
& \underline{0.9643} \\

Lower-body tracking
& 0.8717
& 0.7522
& \textbf{0.9081}
& \underline{0.9040}
& 0.8593 \\

\midrule

Critical Force --- Forward (N)
& 80
& \underline{190}
& 180
& \textbf{$>200$}
& \underline{190} \\

Critical Force --- Lateral (N)
& 100
& \underline{120}
& 70
& \textbf{$>200$}
& \underline{120} \\

Perturbation Success ($n$ / 40)
& 17 / 40
& \underline{30 / 40}
& 23 / 40
& \textbf{40 / 40}
& \underline{30 / 40} \\

Perturbation Success Rate (\%)
& 42.5
& \underline{75.0}
& 57.5
& \textbf{100.0}
& \underline{75.0} \\

\bottomrule
\end{tabular}

\end{threeparttable}
\end{table*}

\subsection{Simulation Experiments}
\subsubsection{Long-Horizon Switching and Generalization}
\rev{We first evaluate \Method{} on sequential multi-motion tasks to address Q1 and Q2.
We report success rate, cumulative reward using the same reward function as OpHRL, and the mean number of controller switches per episode.
An episode is successful if the entire motion sequence is completed without falling.
The comparison includes the following methods and ablations:}

\begin{itemize}
\setlength{\itemsep}{0pt}
\setlength{\parsep}{0pt}
\setlength{\topsep}{2pt}
    \item \textit{Ablations:}
    Standalone OpVQ-VAE (Sec.~\ref{sec:vqvae_switching}) and OpHRL (Sec.~\ref{sec:ophrl}) evaluate the two switching predictors individually. VQ-VAE removes the option-prediction head during
    representation learning. Continuous Op-AE replaces the discrete tokens in OpVQ-VAE with continuous latent representations, keeping all other components unchanged. BC omits RL fine-tuning, while Vanilla HRL omits option-aware guidance.
    \item \textit{Baselines:}
    Learned Gate, MoE, and Causal Transformer use motion references as input and are trained with supervision from $\mathcal{D}_{\mathrm{Op}}$. Learned Gate predicts the controller ID and uses the same handoff mechanism as \Method{}. MoE blends the two controllers' outputs using learned mixture weights, while Causal Transformer predicts controller selection with a GPT-style autoregressive architecture~\cite{radford2018improving}. Option-Critic~\cite{Bacon2017OptionCritic} learns option selection and termination through policy-gradient optimization.
    \item \textit{Fixed policies:} Decoupled and Coupled use one controller throughout each sequence.
    \item \textit{Offline oracle:}
    $\mathcal D_{\mathrm{Op}}$ and $\mathcal D_{\mathrm{trans}}$ oracles directly replay recorded switch timings, pre-evaluated for each motion chain through offline rollouts and stored as lookup tables.
\end{itemize}

\noindent\rev{\textbf{Results.}
As shown in Table~\ref{tab:performance_all_splits}, \Method{} achieves the highest
success rate among deployable methods across all three evaluation splits.
OpHRL performs better than OpVQ-VAE on Train and Test-C, whereas OpVQ-VAE generalizes better to unseen motions in Test-U.
Combining these complementary strengths, \Method{} exceeds both predictors in success rate across all splits. It also achieves the highest Train reward and ties with OpVQ-VAE on Test-U, although its Test-C reward is lower.}

\rev{The representation ablations highlight the value of option-aware discrete tokens for unseen motions. On Test-U, OpVQ-VAE achieves $70.6\%$ success, compared with $49.4\%$ for VQ-VAE and $61.9\%$ for Continuous Op-AE. These results support the use of option-aware learning
and discrete tokenization for generalization beyond the training motion repertoire.}

\rev{\Method{} achieves higher success and reward than
all four switching baselines and both fixed controllers
across all three splits.
These results support combining complementary
switching predictors and adapting controller selection
to changing motion demands.
The poor performance of Vanilla HRL further highlights
the importance of option-aware guidance.}

\rev{\subsubsection{Single-Motion Control and Perturbation Robustness}
We next evaluate single-motion execution to address Q3, comparing success, tracking quality, cumulative reward, and perturbation robustness across diverse motions.}

\noindent\textbf{Baselines.}
We compare \Method{} with recent whole-body controllers: GMT~\cite{chen2025gmt}, FALCON~\cite{zhang2025falcon}, TWIST~\cite{ze2025twist}, and SONIC~\cite{luo2025sonic}. FALCON represents modular decoupled control, whereas
GMT, TWIST, and SONIC employ coupled whole-body motion imitation.

\begin{figure}[t]
    \centering
    \captionsetup[subfigure]{skip=1pt}

    \begin{subfigure}[t]{0.24\textwidth}
        \centering
        \includegraphics[width=\linewidth]{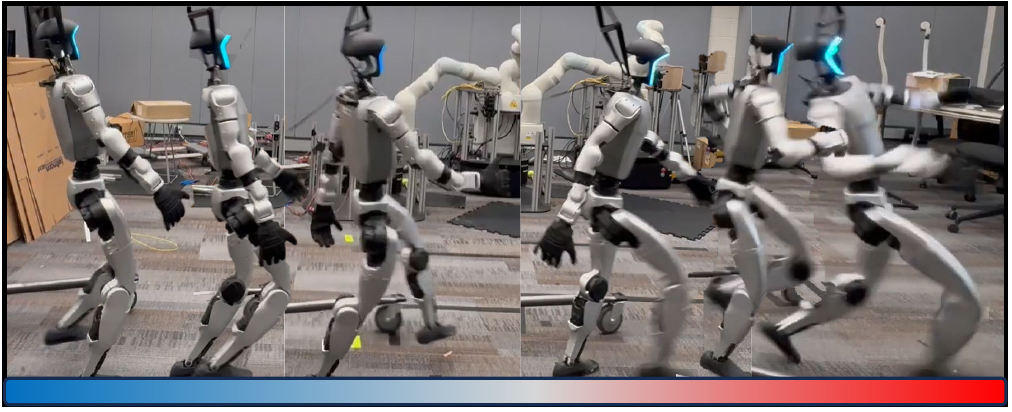}
        \caption{Walking $\rightarrow$ running}
        \label{fig:hw_exp1}
    \end{subfigure}%
    \hspace{0.006\textwidth}%
    \begin{subfigure}[t]{0.24\textwidth}
        \centering
        \includegraphics[width=\linewidth]{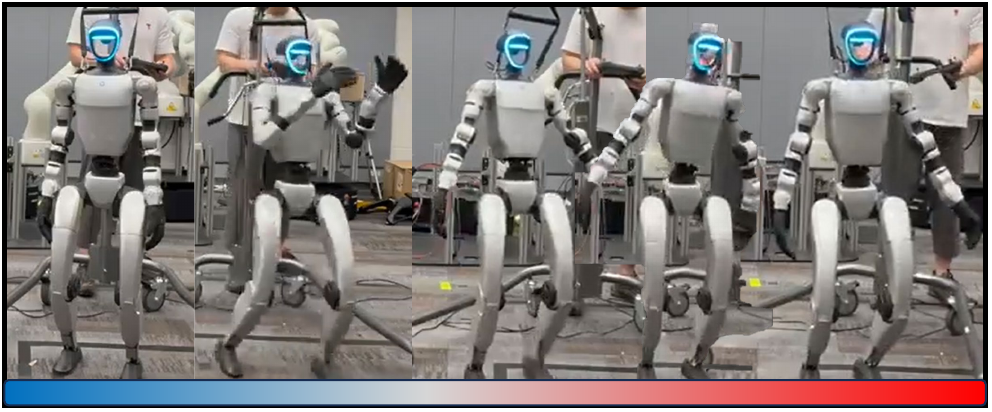}
        \caption{Arm motion $\rightarrow$ jumping}
        \label{fig:hw_exp2}
    \end{subfigure}%
    \hspace{0.006\textwidth}%
    \begin{subfigure}[t]{0.24\textwidth}
        \centering
        \includegraphics[width=\linewidth]{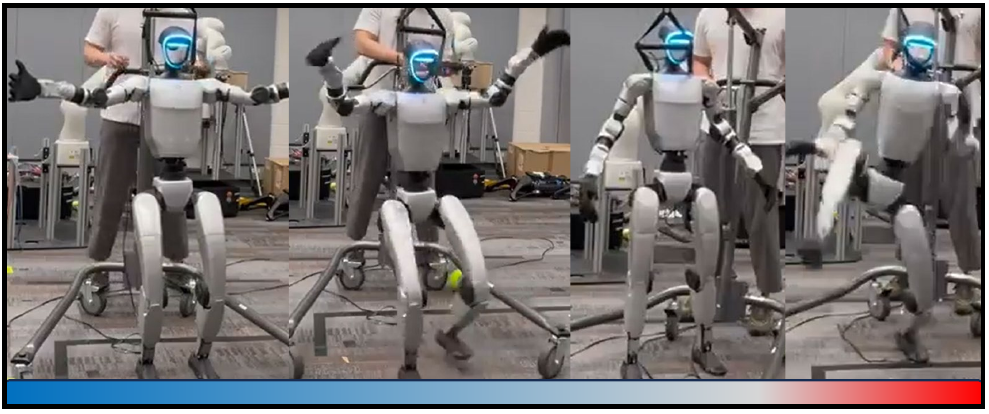}
        \caption{Arm swinging $\rightarrow$ kicking}
        \label{fig:hw_exp3}
    \end{subfigure}%
    \hspace{0.006\textwidth}%
    \begin{subfigure}[t]{0.24\textwidth}
        \centering
        \includegraphics[width=\linewidth]{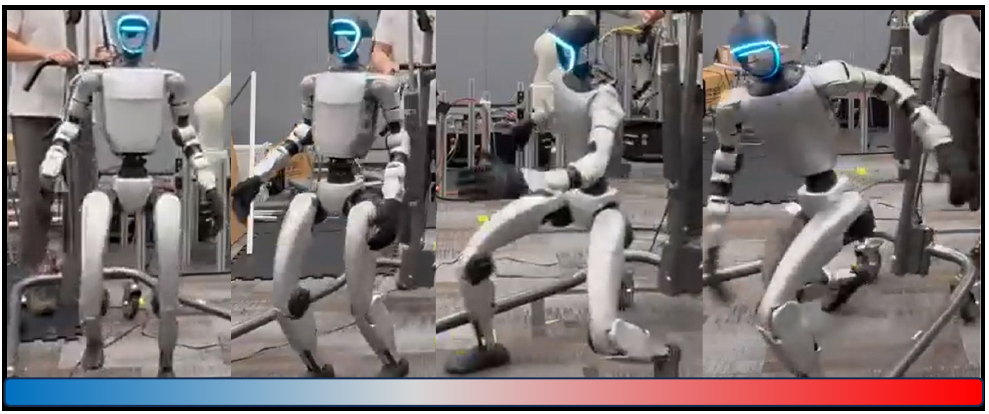}
        \caption{Walking $\rightarrow$ crouched walk}
        \label{fig:hw_exp4}
    \end{subfigure}

    \begin{subfigure}[t]{0.485\textwidth}
        \includegraphics[width=\linewidth]{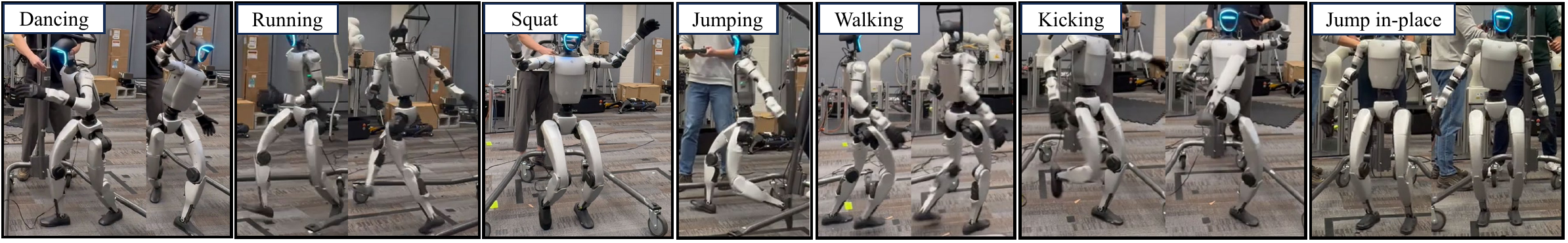}
        \caption{Single-motion execution}
        \label{fig:hw_single_motion}
    \end{subfigure}

    \caption{
    Hardware deployment of \Method{} on the Unitree G1 humanoid.
    (a)--(d) Long-horizon sequences involving online switching between
    the decoupled policy $\pi_D$ (blue) and coupled policy $\pi_C$ (red). (e) Single-motion tasks with adaptive policy selection, where \Method{} automatically selects the more suitable low-level policy based on motion characteristics.}
    \label{fig:hw_results}
    \vspace{-0.5cm}
\end{figure}

\noindent\rev{\textbf{Results.}
As shown in Table~\ref{tab:method_comparison}, \Method{} achieves the highest success rate and mean reward. Since the mean reward jointly captures tracking, stability, and motion smoothness, this result demonstrates that \Method{} achieves a strong balance
among these objectives. The fixed policies remain strongest in their specialized regimes: FALCON emphasizes precise upper-body control, whereas coupled policies provide stronger lower-body tracking during dynamic motions. Although \Method{} does not maximize every body-specific tracking metric independently, it avoids the sharp degradation of either specialist outside its preferred regime and maintains a more balanced overall performance through adaptive selection. Under external perturbations, \Method{} matches FALCON for second-best robustness behind SONIC, while retaining the highest overall success and reward. }

\subsection{Hardware Experiment}
To validate the effectiveness of \Method{} on real hardware (Q4), we deploy our method on the G1 humanoid platform. As shown in Fig. \ref{fig:hw_results}, our approach enables successful execution of sequential tasks by switching between policies, allowing both dynamic and static motions to be performed within a unified framework.
\rev{The hardware experiments cover diverse individual motions and
sequences involving switching between different motions
(see the supplementary video).
We observed consistent switching across three repeated trials for each hardware demonstration. These results are qualitative demonstrations on a selected subset of motions, rather than a systematic hardware success-rate evaluation over the full repertoire.}

\rev{On hardware, motions with substantial low-level tracking degradation can destabilize switching by shifting OpHRL's proprioceptive inputs. Distilling it into a reference-only switching policy may reduce this sensitivity. }
\section{conclusion}
\rev{We present \Method{}, an online policy-switching framework that combines coupled and decoupled controllers for whole-body humanoid control. 
OpHRL uses cross-simulator switch-boundary guidance and closed-loop reinforcement learning to improve switching over BC and vanilla HRL, while OpVQ-VAE provides complementary generalization through option-aware discrete motion tokens. 
A token-familiarity router selects between the two predictors according to the familiarity of the current motion-token sequence. 
Simulation results show that \Method{} improves long-horizon task success by combining precise, robust decoupled control with agile whole-body control, and qualitative hardware experiments demonstrate zero-shot deployment on the Unitree G1.}

\rev{However, the framework lacks explicit environment and object-interaction awareness (e.g., perturbations, uneven terrain, or contact-rich manipulation), suggesting future work on perceptive and interaction-aware control. \Method{} is modular and can further benefit from stronger components.}


\bibliographystyle{IEEEtran}
\bibliography{reference}


\end{document}